\documentclass{article}

\usepackage{microtype}
\usepackage{graphicx}
\usepackage{subfigure}
\usepackage{booktabs} % for professional tables
\usepackage{adjustbox}

\usepackage{hyperref}

\usepackage[accepted]{icml2019}

\icmltitlerunning{Multivariate Time Series Classification using Dilated Convolutional Neural Network}

\begin{document}

\twocolumn[
\icmltitle{Multivariate Time Series Classification using Dilated Convolutional Neural Network}



%\icmlsetsymbol{*}

\begin{icmlauthorlist}
\icmlauthor{Omolbanin Yazdanbakhsh}{xsensor}
\icmlauthor{Scott Dick}{uofa}
\end{icmlauthorlist}

\icmlaffiliation{uofa}{Department of Electrical and Computer Engineering, University of Alberta, Edmonton, Canada}
\icmlaffiliation{xsensor}{Xsensor Technology Corp, Calgary, Alberta, Canada}

\icmlcorrespondingauthor{Omolbanin Yazdanbakhsh}{yazdanba@ualberta.ca}
\icmlcorrespondingauthor{Scott Dick}{scott.dick@ualberta.ca}

\icmlkeywords{Multivariate, Deep Learning, Time Series}

\vskip 0.3in
]



\printAffiliationsAndNotice{}  % leave blank if no need to mention equal contribution

\begin{abstract}
Multivariate time series classification is a high value and well-known problem in machine learning community. Feature extraction is a main step in classification tasks. Traditional approaches employ hand-crafted features for classification while convolutional neural networks (CNN) are able to extract features automatically. In this paper, we use dilated convolutional neural network for multivariate time series classification. To deploy dilated CNN, a multivariate time series is transformed into an image-like style and stacks of dilated and strided convolutions are applied to extract in and between features of variates in time series simultaneously. We evaluate our model on two human activity recognition time series, finding that the automatic features extracted for the time series can be as effective as hand-crafted features.
\end{abstract}

\section{Introduction}
\label{intro}

Time series are observations recorded sequentially over time. In univariate time series, data are collected from one source, thus, each observation is a single scalar while in multivariate time series, observations are recorded from multiple sources simultaneously, thus, each data point is a multi-dimensional vector. Time series applications arises in a broad range of domains including health care \cite{Jones2009}, climate \cite{Zhang2018}, robotics \cite{perez2015fast} and stock markets \cite{maknickiene2011investigation}. 

Various statistical and machine learning algorithms have been proposed for time series classification \cite{Yazdanbakhsh2017,Che2018,Saad2017,Dobigeon2007,Zhang2003}. General approach for time series classification is splitting time series to equal size segments using a fixed-length sliding window and extracting hand-crafted features from the segments for classification tasks. The features are usually statistical measurements or features extracted from another domain such Fourier and Wavelet domain \cite{jiang2015,Ravi2017,lin2003}. In multivariate time series classification, commonly, information is extracted separately from each variate, and the features are concatenated for the classification task \cite{Zheng2016,Liu2019,Chambon2018}. However, it may ignore the relation between data items comprising each component of an observation. 

Deep learning approaches substitute hand-crafted features with features learned automatically from data. In CNN, the features are obtained by learning filters convolved with small sub-region of data \cite{krizhevsky2012,LeCun1998}. Performance of a wide range of applications have been improved by replacing traditional approaches with deep learning methods \cite{van2016,toshev2014deeppose,Young2018}. Various deep learning methods have been employed for time series classification \cite{Zheng2016,Liu2019,Gao2018}. However, in most studies, hand-crafted features are still injected to the network to improve its performance \cite{Ravi2017,Ignatov2018}. 

In this paper, we propose a convolution-based learning algorithm for multivariate time series classification. In this algorithm, a multivariate time series is studied as a one-channel image where each row in the image corresponds to one of the variates in the time series. The CNN employed in this work is inspired by Wavenet architecture \cite{van2016} which applied dilated convolution operations for audio synthesis. We show that this structure is able to learn the relation between variates in the time series and uses their mutual information to reduce noise and improve performance. This method is applied to two human activity recognition time series (WISDM v.1.1 \cite{Kwapisz2011} and WISDM v.2 \cite{lockhart2011design}) and its performance is compared against published results on these time series. 

The remainder of the paper is organized as follow. In Section ~\ref{relat}, we review related works on time series studies using deep learning methods. Our method is presented in Section ~\ref{method} and our experimental results are presented in Section ~\ref{experim}. We close with a summary in Section ~\ref{colnc}.

\section{Related Works}
\label{relat}

Various approaches have been proposed for time series analysis. \cite{Zhang2003} combined auto regressive integrated moving average model (ARIMA) and neural network models for time series prediction. \cite{Yazdanbakhsh2017} used neuro-complex fuzzy systems for multivariate time series prediction. \cite{Che2018} applied an extension of recurrent neural network (RNN) for multivariate time series classification with missing values. \cite{yeo2017} employed long short-term memory (LSTM) model for chaotic time series forecasting. \cite{Orozco2018} developed an ordinal regression deep neural network based on LSTM. \cite{Saad2017} proposed a Bayesian non-parametric method for multivariate time series forecasting. 

Convolution-based approaches have been proposed for time series analysis as well. \cite{Borovykh2017} uses WaveNet model for conditional multivariate time series prediction where forecasting of each variate is conditioned on other variates in the time series. \cite{Zheng2016} designed a multi-channel CNN where each channel takes a variate of multivariate time series as input and learns features individually; classification task is done by applying a multi-layer perceptron network on the combined features of each channel. \cite{Lee2017} applied a CNN for fault classification and diagnostic in semiconductor manufacturing where the relation between variates are only explored over time. \cite{Liu2019} did multivariate time series classification in a four-stage process; the time series is converted to a 3-D tensor and passed through three stages including univariate convolution stage, multivariate convolution stage and fully connected stage. \cite{Gao2018} studied multivariate time series by partitioning the data to groups based on covariance structure of the time series. \cite{Chambon2018} employed deep learning for sleep stage classification; features are extracted separately from each channel of Polysomnography (PSG) signals and combined to give a classification label. \cite{Gamboa2017} did a review on deep learning approaches for time series analysis. \cite{fawaz2019deep} benchmarked some well-known deep learning algorithms for univariate and multivariate time series classification.

Moreover, WISDM time series has been studied using deep learning approaches. \cite{Ravi2017} applied deep learning algorithms on spectrogram features of the time series. \cite{Ignatov2018} extracted features from each variate of the time series separately using CNN; more over, statistical features are added to the network as additional information.

\section{Methods}
\label{method}

Our proposed system is shown in Figure 1. It has three components a) image module 2) CNN module 3) fully connected module. Detailed description of each module is as follow:

% Could change the width to columnwidth instead.

\begin{figure*}[ht]
\vskip 0.2in
\begin{center}
\centerline{\includegraphics[width=\textwidth]{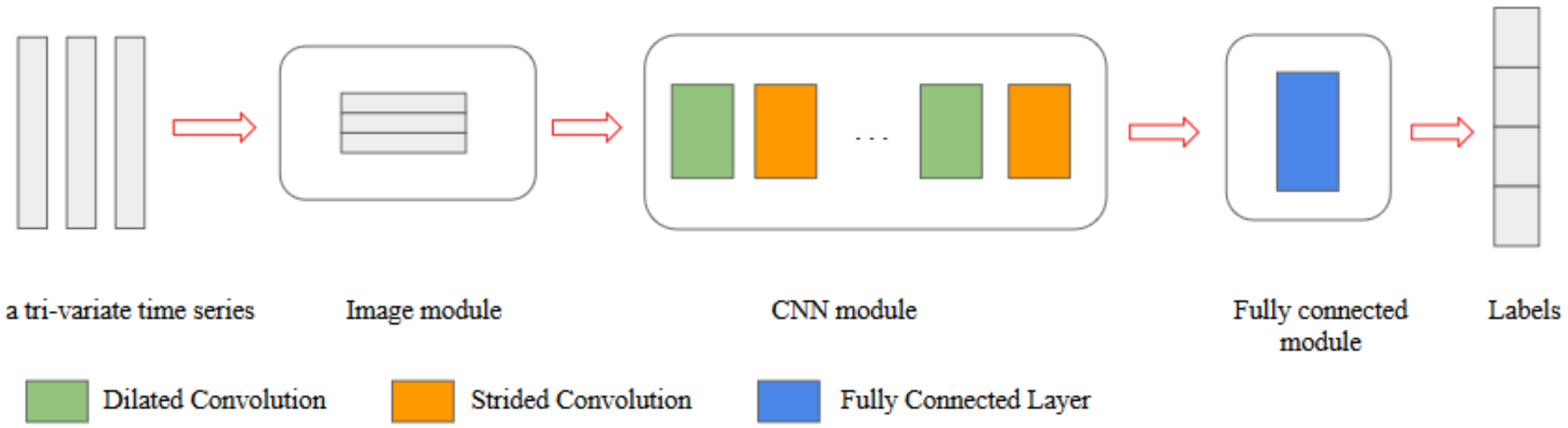}}
\caption{Block diagram of the proposed approach}
\label{block-diagram}
\end{center}
\vskip -0.2in
\end{figure*}

\subsection{Image Module}
In this module, first, multivariate time series is split to equal-size segments. A sliding window is moved over the time series to give batches of the time series; selecting between overlapping and non-overlapping sliding window is application-dependent. Each segment is a multi-dimensional vector of size ($K \times M$) where $K$ is length of the sliding window and $M$ is number of variates in the time series.
Given the segments of time series, we transform each segment to a one-channel image where each variate in the segment forms a row in the image. Outputs of this module are one-channel images of size ($M \times K$).

\subsection{CNN Module}
CNN module comprises of stacks of dilated convolution and strided convolution layers with ReLU activations. Dilated convolution layers are applied to extract features from the segments of multivariate time series obtained from the image module, and strided convolution layers reduce spatial dimension. The first layer in the CNN module is a dilated convolution layer; the filters in this layer slide over a one-channel image of size ($M \times K$)  with stride=1 in both horizontal and vertical direction; sliding the filters in the both directions extracts features in and between variates simultaneously, thus, there is no need to study each variate individually at the beginning of our process. Moreover, employing dilation rate in the convolution allows the model to learn relation between observations that are far apart.

Next layers are stacks of strided and dilated convolution layers processing the obtained feature maps. Filters in the strided convolution layers are applied on each row separately, thus, only reduce number of columns of the feature maps and preserve number of rows. 

\subsection{Fully Connected Module}

In this module, fully connected layers are applied to the output of the CNN module, then, a Softmax layer computes probability of predicted classes. As loss function, cross-entropy function is used.

\section{Experiments}
\label{experim}

To evaluate the performance of the proposed model, we apply the model to two public multivariate time series containing accelerometer data. The results are compared against published results on these datasets.

\subsection{Time Series}

WISDM v1.1 \cite{Kwapisz2011} has accelerometer data (x-, y-, z- channels) collected from 36 users carrying an Android phone over their front pocket while performing a set of daily activities including walking, jogging, upstairs, downstairs, sitting, and standing. Table ~\ref{wisdm-v11} shows details of this data set.

\begin{table*}[tb]
\caption{Details of WISDM v.1.1}
\label{wisdm-v11}
\vskip 0.15in
\begin{center}
\begin{small}
\begin{sc}
\begin{adjustbox}{width=1\textwidth}
\begin{tabular}{lcccccr}
\toprule
%Data set & Naive & Flexible & Better? & Worse \\
Activity&Walking&Jogging&Upstairs &Downstairs & Sitting & Standing\\
\midrule
Number of Examples & 424,400 (38.6\%) &	342,177 (31.2\%) & 122,869 (11.2\%) & 100,427 (9.1\%) & 59,939 (5.5\%) & 48,395 (4.4\%) \\
\multicolumn{7}{l}{Total number of examples= 1,098,207} \\
\multicolumn{7}{l}{Number of activities= 6} \\
\multicolumn{7}{l}{Sampling rate= 20 Hz} \\
\bottomrule
\end{tabular}
\end{adjustbox}
\end{sc}
\end{small}
\end{center}
\vskip -0.1in
\end{table*}

We create two data sets from this time series in order to be able to compare the results with published results. The first data set (v1-split) has non-overlapping segments with length 100 (5 sec) and the segments are split to 80\% and 20\% as training and testing sets, respectively, giving 8347 instances as training and 2643 instances as testing. The second data set (v1-individual) considers data of 28 users as training set and 8 users as training; the training and testing sets are segmented with sliding window of 200 (10 sec) with step size of 20 giving 41729 training and 13162 testing instances.

WISDM v.2 \cite{lockhart2011design} has x-, y-, z- channels of accelerometer collected from 536 users while walking, jogging, stairs, sitting, standing and lying down. Table ~\ref{wisdm-v2} shows details of this time series.

\begin{table*}[tb]
\caption{Details of WISDM v.2}
\label{wisdm-v2}
\vskip 0.15in
\begin{center}
\begin{small}
\begin{sc}
\begin{adjustbox}{width=1\textwidth}
\begin{tabular}{lcccccr}
\toprule
%Data set & Naive & Flexible & Better? & Worse \\
Activity&Walking&Jogging&Upstairs &Downstairs & Sitting & Standing\\
\midrule
Number of Examples &1,255,923 (42.1\%)	&438,871 (14.7\%)	&57,425 (1.9\%)	&663,706 (22.3\%)	&288,873 (9.7\%)	&275,967 (9.3\%) \\
\multicolumn{7}{l}{Total number of examples= 2,980,765} \\
\multicolumn{7}{l}{Number of activities= 6} \\
\multicolumn{7}{l}{Sampling rate= 20 Hz} \\
\bottomrule
\end{tabular}
\end{adjustbox}
\end{sc}
\end{small}
\end{center}
\vskip -0.1in
\end{table*}

Non-overlapping segments with length 200 (10 sec) is obtained from the time series; the segments are split to 80\% and 20\% as training and testing sets, respectively, giving 10396 segments as training and 4456 segments as testing.

\subsection{Network Analysis}

The network designed for WISDM v1-individual has six layers including:
DL (filter size= (3,20), number of filters=32, dilation rate=(1,2)) $\rightarrow$ SL (filter size= (1,4), number of filters=32, stride =(1,4)) $\rightarrow$ DL (filter size= (3,3), number of filters=32, dilation rate=(1,2)) $\rightarrow$ SL (filter size= (1,4), number of filters=32, stride =(1,4)) $\rightarrow$ FL (unit=1024) $\rightarrow$ FL (unit=6)
where DL, SL and FL stand for dilated convolution layer, strided convolution layer and fully connected layer, respectively. 

The network designed for WISDM v1-split is as follow:
 DL (filter size= (3,10), number of filters=32, dilation rate=(1,2)) $\rightarrow$ SL (filter size= (1,4), number of filters=32, stride =(1,4)) $\rightarrow$ DL (filter size= (3,3), number of filters=32, dilation rate=(1,2)) $\rightarrow$ SL (filter size= (1,2), number of filters=32, stride =(1,2)) $\rightarrow$ FL (unit=1024) $\rightarrow$ FL (unit=6)

To train the network, we use batch size of 256 on a single GPU. Adam optimizer \cite{kingma2014adam} is used with $\beta_1=0.9$ and $\beta_2=0.999$ and a the learning rate was set to $10^{-4}$. We also apply $L_2$ regularization with weight $10^{-5}$ for v1-individual and $10^{-3}$ for v1-split.

The network designed for WISDM v.2 is trained same as WISDM v.1.1 with  $L_2$ regularization weight of $10^{-5}$; the model has 8 layers as follow:
DL (filter size= (3,10), number of filters=32, dilation rate=(1,2)) $\rightarrow$ SL (filter size= (1,2), number of filters=32, stride =(1,2)) $\rightarrow$ DL (filter size= (3,3), number of filters=32, dilation rate=(1,2)) $\rightarrow$ SL (filter size= (1,2), number of filters=32, stride =(1,2)) $\rightarrow$ DL (filter size= (3,3), number of filters=64, dilation rate=(1,1)) $\rightarrow$ SL (filter size= (1,2), number of filters=64, stride =(1,2)) $\rightarrow$ FL (unit=512) $\rightarrow$ FL (unit=6)

Table ~\ref{results-v1-split} shows performance of our network in terms of F1-score \cite{goutte2005} for WISDM v1-split. We compare our results against \cite{Ravi2017} which passes spectrogram features of non-overlapping segments of the time series to a deep learning algorithm. 

\begin{table}[tb]
\caption{Accuracy results for WISDM v1-split based on F1-score}
\label{results-v1-split}
\vskip 0.15in
\begin{center}
\begin{small}
\begin{sc}
%\begin{adjustbox}{width=1\textwidth}
\begin{tabular}{lcc}
\toprule
%Data set & Naive & Flexible & Better? & Worse \\
 &Our Method & \cite{Ravi2017} \\
\midrule
Walking	&97.4\%	&99.3\% \\
Jogging	&98.3\%	&99.5\% \\
Upstairs	&86.4\%	&95.3\% \\
Downstairs	&80.5\%	&95.1\% \\
Sitting	  	&98\%	&98.2\% \\
Standing	&94.9\%	&97.6\% \\
\bottomrule
\end{tabular}
%\end{adjustbox}
\end{sc}
\end{small}
\end{center}
\vskip -0.1in
\end{table}

Table ~\ref{results-v1-indv} shows our results for WISDM v1-individual based on recall \cite{goutte2005} and compares it to \cite{Ignatov2018} which creates segments in training and testing sets based on individuals;  the network uses a CNN to extract features from each variate separately and combines them with statistical features for classification. Table ~\ref{results-v2} shows performance of our method on WISDM v.2 based on F1-score and compares it with \cite{Ravi2017}.

\begin{table}[tb]
\caption{Accuracy results for WISDM v.1 individual based on recall}
\label{results-v1-indv}
\vskip 0.15in
\begin{center}
\begin{small}
\begin{sc}
%\begin{adjustbox}{width=1\textwidth}
\begin{tabular}{lcc}
\toprule
%Data set & Naive & Flexible & Better? & Worse \\
 &Our Method & \cite{Ignatov2018} \\
\midrule
Walking	&97.8\%	&97.8\% \\
Jogging	&95.5\%	&98.5\% \\
Upstairs	&78.5\%	&72.2\% \\
Downstairs	&75.1\%	&87.0\% \\
Sitting		&90.9\%	&82.6\% \\
Standing	&91.8\%	&93.3\% \\
\bottomrule
\end{tabular}
%\end{adjustbox}
\end{sc}
\end{small}
\end{center}
\vskip -0.1in
\end{table}

\begin{table}[tb]
\caption{Accuracy results for WISDM v.2 based on F1-score}
\label{results-v2}
\vskip 0.15in
\begin{center}
\begin{small}
\begin{sc}
%\begin{adjustbox}{width=1\textwidth}
\begin{tabular}{lccccccr}
\toprule
%Data set & Naive & Flexible & Better? & Worse \\
 &Our Method & \cite{ Ravi2017} \\
\midrule
Walking	&96.6\%	&97.2\% \\
Jogging	&96.9\%	&97.9\% \\
Stairs		&63.1\%	&79.3\% \\
Sitting		&91.2\%	&88.2\% \\
Standing	&87.2\%	&82.1\% \\
Lying Down	&90.7\%	&87.2\% \\
\bottomrule
\end{tabular}
%\end{adjustbox}
\end{sc}
\end{small}
\end{center}
\vskip -0.1in
\end{table}

Table ~\ref{results-v1-split} and ~\ref{results-v2}  indicate that features extracted automatically using our model can be as effective as spectrogram features. In WISDM v.2, for walking and jogging activities, our accuracy is 1\% lower and our model outperforms  \cite{ Ravi2017}  in the static activities including sitting, standing and lying down (3.9\% higher on average). However, the model cannot compete in the stairs activity where we have the fewest number of instances (1.9\%). In WISDM v1-split,  \cite{ Ravi2017}] outperforms our model in all activities; our accuracy is lower 8.9\% and 14.6\% for upstairs and downstairs, respectively and for the other activities, our accuracy is 1.5\%  lower on average. Table ~\ref{results-v1-indv} indicates, our model outperforms combination of statistic features and CNN features \cite{Ignatov2018} for upstairs and standing activities (2.4\% higher on average), and has similar performance on walking. For jogging, downstairs and standing is lower 5.5\% on average.

\section{Conclusion}
\label{colnc}

We have proposed a new algorithm for multivariate time series classification. After splitting a multivariate time series to equal size segments, we convert the segments to one-channel images; the one-channel images are processed using stacks of dilated and strided convolutions. The extracted features for classification task are obtained by considering the inter and intra relation between variates. Our experiments show that the proposed model can be as effective as models working with hand-crafted features such as spectrogram and statistical features.


\bibliography{multivariateTimeseries2}
\bibliographystyle{icml2019}

\end{document}